\documentclass[sigconf]{acmart}

\AtBeginDocument{%
  \providecommand\BibTeX{{%
    \normalfont B\kern-0.5em{\scshape i\kern-0.25em b}\kern-0.8em\TeX}}}

\setcopyright{acmcopyright}
\copyrightyear{2018}
\acmYear{2018}
\acmDOI{10.1145/1122445.1122456}

\acmConference[Seattle '20]{Seattle '20: ACM MULTIMEDIA}{October 12--16 , 2020}{Seattle, USA}
\acmBooktitle{Seattle '20: ACM MULTIMEDIA,
  October 12--16 , 2020, Seattle, USA}
\acmPrice{15.00}
\acmISBN{978-1-4503-XXXX-X/18/06}
\usepackage{url}
\usepackage{multirow}

\begin{document}

\title{Multi-view Drone-based Geo-localization via Style and Spatial Alignment}

\author{Siyi Hu}
\email{siyi.hu@monash.edu}
\orcid{1234-5678-9012}
\affiliation{%
  \institution{Faculty of Information and Technology, Monash University}
  \streetaddress{Wellington Road}
  \city{Melbourne}
  \state{VIC}
  \postcode{3800}
}

\author{Xiaojun Chang}
\email{xiaojun.chang@monash.edu}
\affiliation{%
  \institution{Faculty of Information and Technology, Monash University}
  \streetaddress{Wellington Road}
  \city{Melbourne}
  \state{VIC}
  \postcode{3800}
}

\renewcommand{\shortauthors}{Trovato and Tobin, et al.}


\begin{abstract}
  In this paper, we focus on the task of multi-view multi-source geo-localization, which serves as an important auxiliary method of GPS positioning by matching drone-view image and satellite-view image with pre-annotated GPS tag. To solve this problem, most existing methods adopt metric loss with an weighted classification block to force the generation of common feature space shared by different view points and view sources. However, these methods fail to pay sufficient attention to spatial information (especially viewpoint variances). To address this drawback, we propose an elegant orientation-based method to align the patterns and introduce a new branch to extract aligned partial feature. Moreover, we provide a style alignment strategy to reduce the variance in image style and enhance the feature unification. To demonstrate the performance of the proposed approach, we conduct extensive experiments on the large-scale benchmark dataset. The experimental results confirm the superiority of the proposed approach compared to state-of-the-art alternatives.

\end{abstract}
% \begin{figure*}[t]
% \begin{center}
%   \includegraphics[width=0.9\linewidth]{u-0.jpg}
% \end{center}
%   \caption{Result of color uniform on satellite-view images from University-1652.}
% \label{fig:long}
% \label{fig:onecol}
% \end{figure*}


\begin{teaserfigure}
  \includegraphics[width=\textwidth]{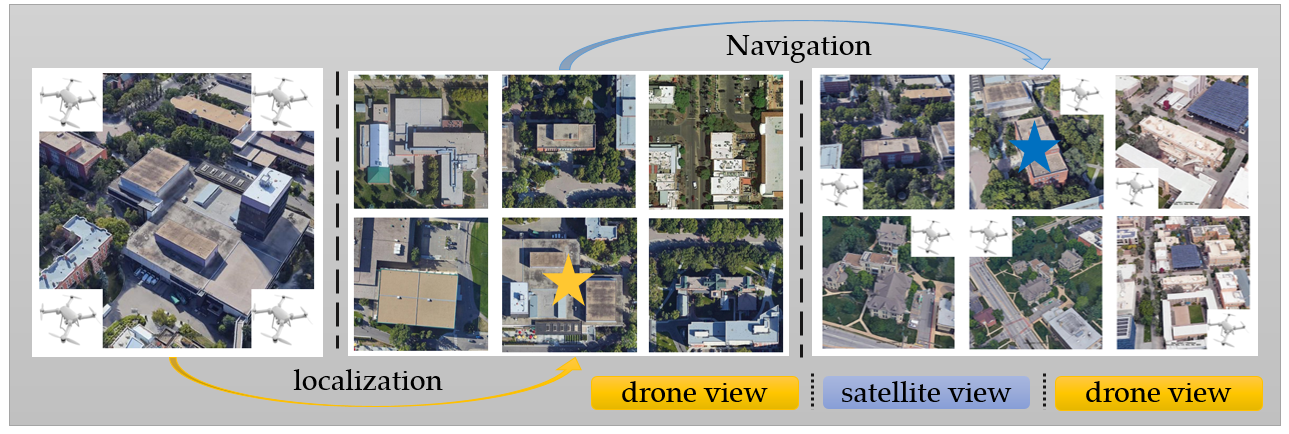}
  \caption{Drone-based geo-localization and navigation. Stars indicate the correct match.}
  \label{fig:teaser}
\end{teaserfigure}

\maketitle

\section{Introduction}
\begin{figure}[t]
\begin{center}
   \includegraphics[width=1.0\linewidth]{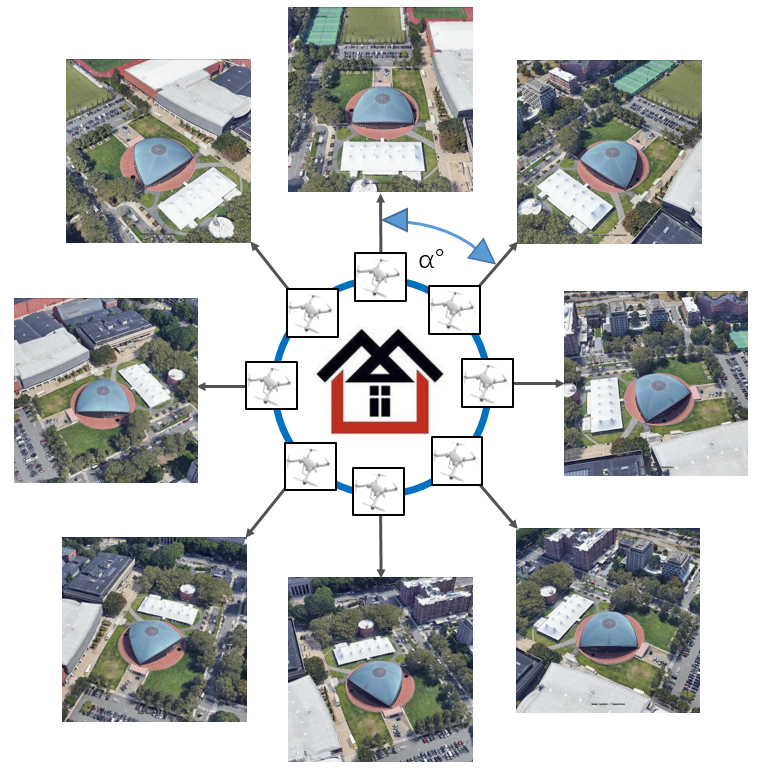}
\end{center}
   \caption{Drone-based geo-localization is a multi-view task. When flying around the building, the drone-view camera can capture rich information about the target, including scale and viewpoint variance.  }
\label{fig:long}
\label{fig:onecol}
\end{figure}

Cross-view geo-localization has attracted increasing attention in the past few years \cite{lin2015learning} \cite{workman2015wide} \cite{vo2016localizing} \cite{tian2017cross} \cite{zhai2017predicting} \cite{hu2018cvm} \cite{liu2019lending} \cite{liu2019stochastic} \cite{shi2019optimal} \cite{vo2016localizing} \cite{kim2017learned} \cite{bansal2014geometric} \cite{jin2015predicting} \cite{hammoud2013overhead} \cite{jin2017learned} \cite{kim2017satellite} \cite{viswanathan2014vision} \cite{zunker2019house} \cite{basedspatial}. This task aims at localizing the target using only images with pre-annotated GPS tags. Given a query image, we can match the paired satellite-view images and use the GPS tag to determine the location of the user (ground-view image). The cross-view image based geo-localization task shows us the offline localization without GPS information is possible when we are able to match images from different views.

% Usually, bird-view or satellite-view images with GPS information tags are collected to form a database. Ground-view images are treated as query images without GPS information. 

% s, which can serve as an auxiliary for geo-localization provided that the ground view image is easy to obtain.

Existing work on this task has followed the traditional approach to supervised deep learning methods\cite{lin2015learning} \cite{workman2015wide} \cite{vo2016localizing} \cite{tian2017cross} \cite{zhai2017predicting}.  The main purpose of these work is to mine the shared features between ground-view and satellite-view images. In this way, the geo-localization task can be defined as a binary classification problem. 

Therefore, Triplet loss \cite{hoffer2015deep} and Siamese architecture \cite{bromley1994signature} has been widely used to handle this task \cite{hu2018cvm} \cite{liu2019lending} \cite{liu2019stochastic} \cite{shi2019optimal} . Based on this, a large number of attention mechanisms have been proposed to improve the feature alignment from different views \cite{liu2019stochastic} \cite{shi2019optimal}. These approaches did not perform well on this task. The main reason for this is obvious: it is difficult even for a human to find the correct match between a given query ground-view image and a target satellite-view image. Some researchers has used orientation information to further improve the model performance\cite{liu2019lending}, however, there is still a gap between these models and real-world use.

% (CVUSA\cite{zhai2017predicting}) and Australia (CVACT\cite{liu2019lending}). These datasets are only slightly different in terms of total number of images. Moreover, CVACT provide additional orientation tags on each ground-view image.
% Large datasets have been built to cover a large areas \cite{liu2019lending} \cite{zhai2017predicting},

In recent years, with the fast development of the map tools like Google Earth and functions provided by Google Maps API \cite{url1} \cite{url2}, multi-view multi-source images with rich geo-information have become available for online collection. Moreover, drone based tasks are becoming more and more important and have come to play the key roles in areas such as agriculture, aerial photography, navigation, event detection and accurate delivery. Drone-view-based geo-localization tasks such as navigation and target localization are also gaining more attention.

With the release of drone-based multi-view image dataset named University-1652\cite{zheng2020university}, the geo-localization task with rich spatial information towards higher accuracy on image-based geo-localization task has become possible.

With different platforms, viewpoints and increase amount of multi-source multi-view data, drone-view based geo-localization tasks is no longer a binary classification problem. The feature extraction can be made more robust and cover more scenarios, which is more practical and valuable. At the same time, ideas and methods from other image retrieval task like person re-identification can be adopted or learned from.

The implementation of the vanilla method on University-1652\cite{zheng2020university} shows the robustness on sub-tasks and other small released dataset \cite{philbin2007object}  \cite{philbin2008lost} \cite{radenovic2018revisiting} compared to existing work such as CVMNet \cite{hu2018cvm}, Orientation\cite{liu2019lending} and other main benchmarks. However, the vanilla method of multi-source multi-view task suffers from low ability to extract spatial information caused by viewpoints variance. Besides, variance in image style including illumination and fuzziness prejudice the feature unification in both the training and testing stages.

% It also achieves state-of-the-art performance on large dataset like CVUSA when compared to existing work such as CVMNet\cite{hu2018cvm}, Orientation\cite{liu2019lending} and other main benchmarks.

To resolve these problems, we adopt three strategies that significantly improve the drone-based geo-localization performance:
\begin{itemize}
\item To handle the variance in the image style, we provide a style alignment strategy to transform the raw image, which helps to enhance the feature unification.
\item To help CNN capture spatial information about the target with its surroundings, we adopt an orientation based method to align the part feature  with a novel crop method.
\item To enhance the feature extraction, we provide a series of partition strategies to extract partial features. Moreover, we analyze the factor of improvement using different partition strategies  .
\end{itemize}

By applying three strategies, we achieve significant performance improvement compared to the vanilla method, which is a large step towards real-world use.

\begin{figure}[t]
\begin{center}
   \includegraphics[width=0.9\linewidth]{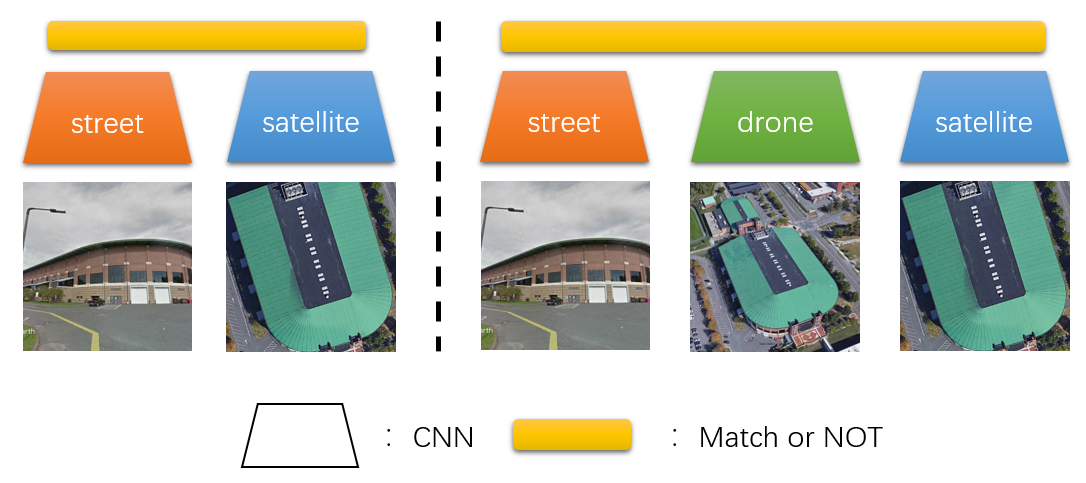}
\end{center}
   \caption{Illustration of the existing geo-localization pipeline (left) and ours (right). Matching between a ground-view image and satellite-view image is difficult even for human. However, drone-view and satellite-view image share more common features with only  viewpoint-caused occlusion and style variance.  }
\label{fig:long}
\label{fig:onecol}
\end{figure}

%-------------------------------------------------------------------------

\begin{figure*}[t]
\begin{center}
   \includegraphics[width=0.9\linewidth]{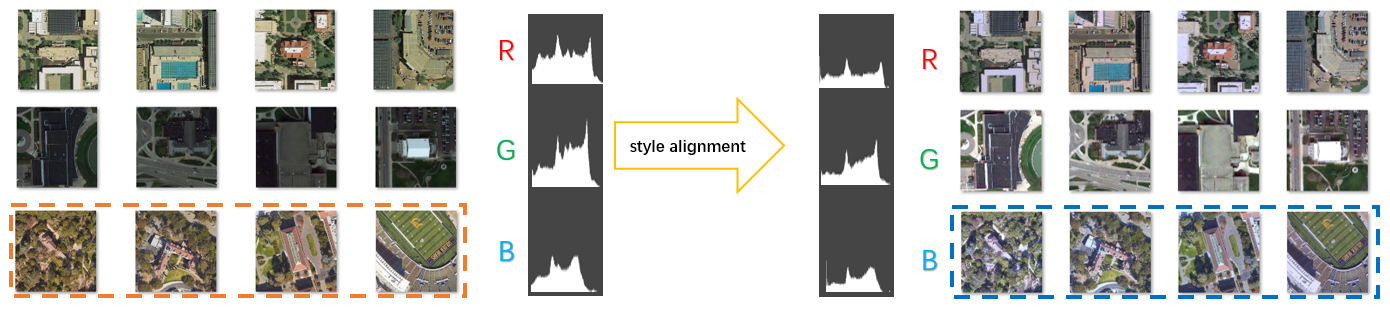}
\end{center}
   \caption{Result of style alignment on satellite-view images from University-1652. The left side presents raw images, while the right side presents images following style alignment. As we can see, the style of raw images with orange dotted frames (left, third row) are warmer than the second row. The red color scale channel is then uniformed to be reduced on amplitude and the style becomes cooler (right, third row). The results show that our method is robust on different styles. }
\label{fig:long}
\label{fig:onecol}
\end{figure*}

\section{Related work}

%-------------------------------------------------------------------------
\subsection{Geo-localization Datasets}

To handle the image based geo-localization problem , several datasets have been built including \cite{lin2015learning}, CVUSA \cite{zhai2017predicting} and CVACT \cite{liu2019lending}. \cite{lin2015learning} was the first well-known cross-view image dataset containing 78k image pairs from two views (i.e. 45° bird view and ground view). Later, CVUSA was released to study the problem of matching the panoramic ground-view image and satellite-view image. CVUSA made the first attempt to conduct user localization when Global Positioning System (GPS) is not available  and serves as an auxiliary localization method. The main difference between CVUSA and \cite{lin2015learning} is that the former focus on localizing the user.  CVACT is another dataset that differs slightly from CVUSA, as CVACT provides user orientation for the ground-view image, which can serve as additional information for better localization performance.

However, these datasets are all paired, which limits the geo-localization problem to a binary classification problem. Worse yet, the viewpoints is fixed in these datasets. In addition, panoramic ground-view image cannot be easily obtained. These disadvantages make it difficult to progress in paired image geo-localization tasks as well as to put them into practical use. 

To relief this limitation, the first drone-based multi-view multi-source image dataset, which was released only recently, is named University-1652 \cite{zheng2020university}. This dataset has three main characteristics, as follows:

\noindent \textbf{Multi-source} University-1652 contains data from three different platforms, namely, satellites, drones and phone cameras. To the best of our knowledge, University-1652 is the first geo-localization dataset to contain drone-view images.

\noindent \textbf{Multi-view} University-1652 contains data from different viewpoints. The ground-view images are collected from different facets of the target buildings. In addition, the synthetic drone-view images capture the target building from various distances and orientations.

\noindent \textbf{More images per class} Unlike the existing datasets that provide image pairs, University-1652 contains 71.64 images per location on average. During training, the increase amount of multi-source multi-view data can help the model to understand the target structure as well as learn the viewpoint-invariant features. 

Compared to CVUSA and CVACT, University-1652 focuses on the relation between images from different sources and different views. Moreover, the task aims at localizing the target in the image (which) rather than localizing the user (where) .

Another contribution of the University-1652 dataset is that it can handle the challenges in real-world drone image collection contexts, considering both the high cost and the privacy and safety issues by using automatically collected synthetic drone-view images. Nevertheless, there is a domain gap between real drone-view image and synthetic drone-view images; the author proves that real drone view images can also work well on models trained by synthetic images \cite{zheng2020university}.

% University-1652 provides street-view, drone-view, satellite-view images along with additional google pictures. The street-view, drone-view and satellite-view images are all collected using Google Earth. It is worth mentioning here that by using the drone-view engine provided by Google Earth, University-1652 has collected multi-viewpoints images of various angles and heights from a spiral curve by simulating a virtual drone. 

\subsection{Geo-localization Method}

\begin{figure*}[t]
\begin{center}
   \includegraphics[width=0.9\linewidth]{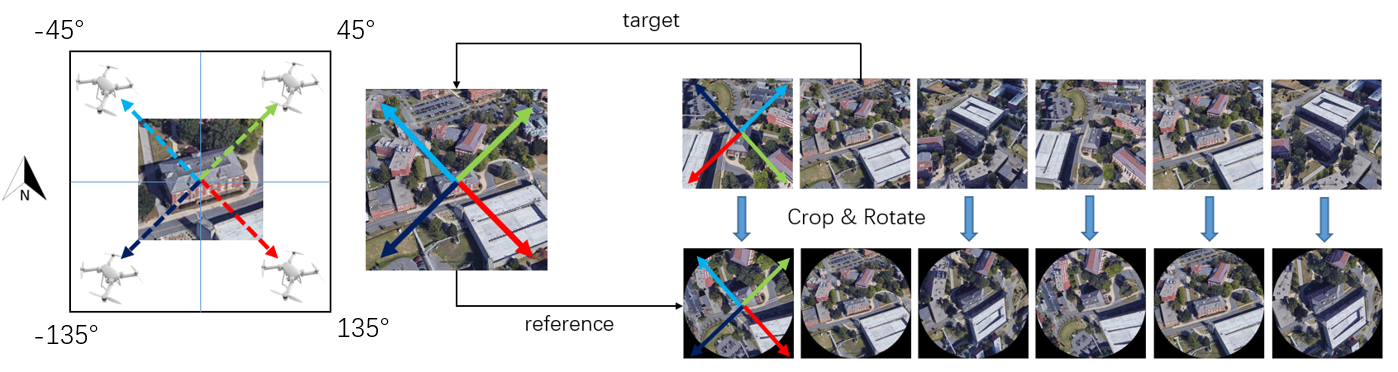}
\end{center}
   \caption{Illustration of 'crop and rotate' image transform method. Picking a raw drone-view image from the first row as a reference, we first crop all images into circles and rotate the other images according to the orientation between itself and the reference image. To ensure compatibility with satellite-view images, we usually adopt the orientation of the satellite as the reference in practice. Colored arrows indicate the orientation reference of the chosen image. After the 'crop and rotate' method, the orientation of other images are aligned. Moreover, the second-row images are somehow more 'similar' to each other than those in the first row, as the surroundings of the target building stay in the same part of the image (for example, the white building with rectangular roof is always in the lower right corner in all six images in the second row).    }
\label{fig:long}
\label{fig:onecol}
\end{figure*}

Most previous works treat the geo-localization as an image retrieval problem \cite{lin2015learning} \cite{workman2015wide} \cite{vo2016localizing} \cite{tian2017cross} \cite{zhai2017predicting} \cite{hu2018cvm} \cite{liu2019lending} \cite{liu2019stochastic} \cite{shi2019optimal}. The key to geo-localization is learning the view-point invariant representation, which aims to bridge the gap between images of different views. Following the development of the deeply learned model, convolutional neural networks (CNNs) have been widely applied to extract the visual features \cite{simonyan2014very} \cite{he2016deep} \cite{ronneberger2015u} \cite{huang2017densely} \cite{xie2017aggregated}. 

One line of works focuses on metric learning and builds a shared space for images collected from different platforms. Workman et al. show that the classification CNN pre-trained on the Place dataset \cite{zhou2017places} can be very discriminative by itself without explicitly fine-tuning \cite{workman2015wide}. The contrastive loss, pulling the distance between positive pairs, could further improve the geo-localization results \cite{lin2015learning}. Recently, Liu et al. propose Stochastic Attraction and Repulsion Embedding (SARE) loss, minimizing the KL divergence between the learned and the actual distributions \cite{liu2019stochastic}. 

Another line of works focus on the spatial misalignment problem in the ground-to-aerial matching. Vo et al. evaluate different network structures and propose an orientation regression loss to train an orientation-aware network \cite{vo2016localizing}. Zhai et al. utilize the semantic segmentation map to improve the semantic alignment  \cite{zhai2017predicting}, while Hu et al. insert the NetVLAD layer \cite{arandjelovic2016netvlad} to extract the discriminative features \cite{hu2018cvm}. Furthermore, Liu et al. propose a Siamese Network to explicitly involve the spatial cues, i.e., orientation maps, into the training \cite{liu2019lending}. Similarly, Shi et al. propose a spatial-aware layer to further improve the localization performance\cite{shi2019optimal}

Following the release of University-1652, since each location has a number of training data points from different views, the model can be trained using a classification CNN with regular cross-entropy loss. The author of university-1652 provides a novel baseline using instance loss to extract the common features from different views and sources. In this way, the viewpoint-invariant feature can be learned in a robust method.

\section{Methods}

\subsection{Style Alignment}

Image-based geo-localization tasks aims at find a robust way of representing the features shared by images of a same place. The learned feature should be invariant despite differences in the viewpoints or sources. Thus an obvious method is to minimize the style invariance both inter-domain and intra-domain.

University-1652 provides us 1652 buildings with various viewpoints and view sources. As the drone-view images are synthesized from the virtual drone engine of Google Earth , the style of drone-view images do not vary significantly. However, the style does vary substantially in satellite-view images. This style variance comes from 1) the satellite-view images being captured in different seasons. (e.g., in winter the image style will be cooler and darker, while in summer and autumn the map style can be warmer and brighter); 2) the satellite-view images being synthesized from multi-color layers with preprocessing.

Existing methods provide us with several kinds of style transfer or style uniform solutions \cite{gatys2016image} \cite{johnson2016perceptual} \cite{li2017universal} \cite{liu2020unity}  \cite{zhu2017unpaired} \cite{huang2017arbitrary}. However, these approaches are all deep learning-based and therefore require a large amount of training data. In our case, there are only 1652 satellite-view images of different university buildings without annotated tags showing which style or class each belongs to. The unsupervised method also does not fit well in this case, as the content of satellite-view images varies widely and the features from the pre-trained model cannot be used directly used without fine-tuning.

Hence, in order to force the satellite-view images to be in a uniform style, we provide a simple color scale-based method that uniforms the image style in a statistical way. Each image has 4 dimension of color scale including: 'S','R','G','B'. The 'S' scale can also be regard as the light scale, as this scale is the main factor in deciding whether a picture is light bias or dark bias. 'R','G','B' represent the red scale, green scale and blue scale, which can decide whether an image is warmer or colder. In addition, some images may have color distortion or color cast caused by an unbalanced color scale. Our style alignment method is simple and efficient for style uniformity on a small-scale dataset without sufficient annotation.

We first compute the mean value of four color scales and record them as the channel mean.

\[{{S\mathop{{}}\nolimits_{{i,j}}}=\frac{{1}}{{3}}{{\mathop{ \sum }\limits_{{i,j \in p}}{R\mathop{{}}\nolimits_{{i,j}}}}+G\mathop{{}}\nolimits_{{i,j}}+B\mathop{{}}\nolimits_{{i,j}}}}\]
\[{S\mathop{{}}\nolimits_{{cm}}=\frac{{1}}{{p}}{\mathop{ \sum }\limits_{{i,j \in p}}{S\mathop{{}}\nolimits_{{i,j}}}}}\]
\[{R,G,B\mathop{{}}\nolimits_{{cm}}=\frac{{1}}{{p}}{\mathop{ \sum }\limits_{{i,j \in p}}{\mathop{{R,G,B}}\nolimits_{{i,j}}}}}\]

Here 'cm' stands for channel mean, while 'p' stands for pixels of the whole image. We then compute the color bias of the different scales and then rescale value of all  channels.

\[{R,G,B\mathop{{}}\nolimits_{{bias}}=R,G,B\mathop{{}}\nolimits_{{cm}}-S\mathop{{}}\nolimits_{{cm}}}\]
\[{scale={{{\mathop{ \sum }\limits_{{i,j \in p}}{S\mathop{{}}\nolimits_{{i,j}}}}}/{S\mathop{{}}\nolimits_{{ave}}}}}\]

% \[{R,G,B\mathop{{}}\nolimits_{{uni}}=R,G,B\mathop{{}}\nolimits_{{raw}} \times scale \times R,G,B\mathop{{}}\nolimits_{{cm}}}\]
We rescale the RGB channel of each pixel by rescale value and uniform the color scale by color bias. Considering that the color scale ranges from 0 to 255, we apply a clip method to prevent color distortion
. 
\[{R,G,B\mathop{{}}\nolimits_{{uni}}=R,G,B\mathop{{}}\nolimits_{{raw}} \times scale \times  \left( 1+{{R,G,B\mathop{{}}\nolimits_{{bias}}}/{S\mathop{{}}\nolimits_{{ave}}}} \right) }\]
\[{R,G,B\mathop{{}}\nolimits_{{final}}=F\mathop{{}}\nolimits_{{clip}} \left( R,G,B\mathop{{}}\nolimits_{{uni}} \right) }\]

We show the result of style alignment in figure 4. In our experiments, we will see the performance improvement obtained by using style alignment strategy. Moreover, our style alignment strategy can be applied to single test image with no prerequisites.

\subsection{Feature Alignment with Partial Feature Extraction}
\begin{figure*}[t]
\begin{center}
   \includegraphics[width=0.9\linewidth]{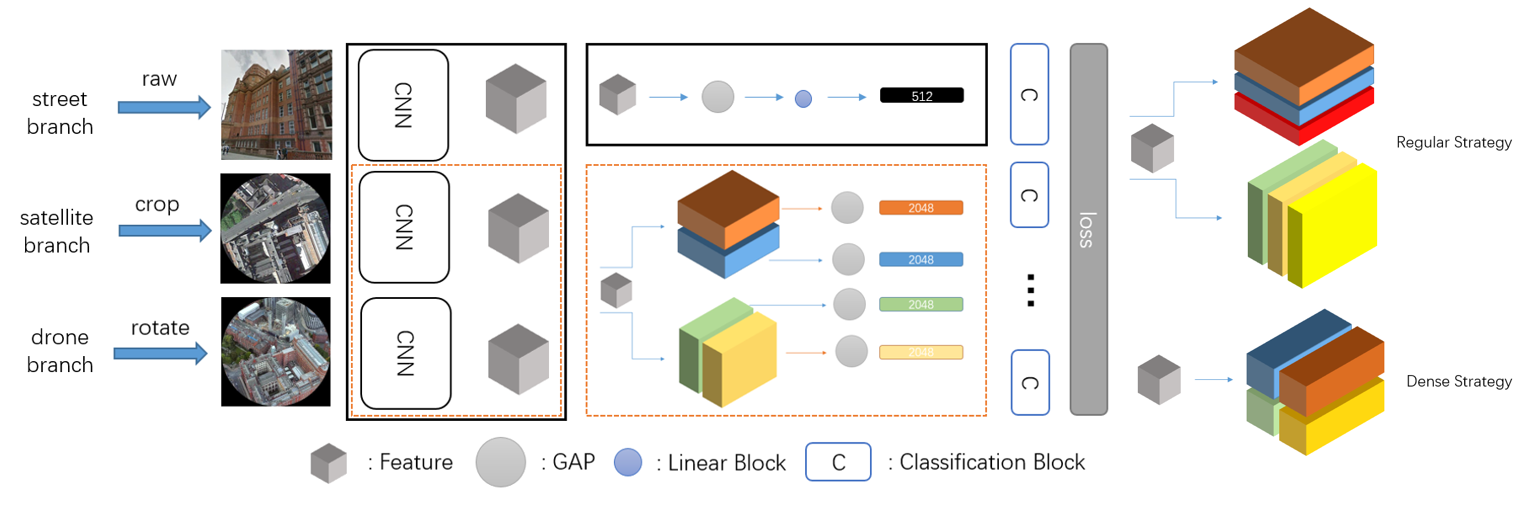}
\end{center}
   \caption{Illustration of our pipelines. Three different sources of images are fed into the model with different data preprocessing strategies. The gray cube represents a feature after the pooling layer of the CNN backbone. The features of satellite branch and drone branch will be fed into the partial feature model (orange dotted frame) followed by a partition strategy. Features of all three branches will be fed into the global feature model (black line frame). The weight of classification block is shared by the global feature model to generate a common space. In practice, considering that there is only one satellite view image per building, we share the weight of the CNN for the satellite branch and drone branch. We show the two partition strategies on the right side of the  pipelines. }
\label{fig:long}
\label{fig:onecol}
\end{figure*}

University-1652 provides us with a vanilla multi-branch model that uses instance loss to guide the CNN in learning the common feature shared by different views. The vanilla model has two main disadvantages: 1) it only focuses on global featurs and 2) it ignores the spatial misalignment according to the variant viewpoint. It can only tell whether the given image contains target building with corresponding surroundings; however, it is not able to distinguish between the relative locations among these buildings due to viewpoint variance. 

A great model should capture the information of relative locations among the target building and its surroundings. Luckily, the drone-view image from  University-1652 are collected following a constant angle step of about 20°. Thus, the most efficient way to get the features aligned is to rotate the raw image. Moreover, the satellite-view images are all captured in the same orientation, which means the feature between the satellite-view and drone-view images can also be aligned by simply rotating a certain number of degree according to the image index.

In order to rotate the images smoothly, we crop the raw image in a novel way. A typical description of a target building's surroundings will resemble e.g. 'within a hundred meters, there is a hospital to the east and a school to the west.' Thus, we only consider information within a circle. We then mask the raw image with a circle such that the radius equals half of the side length.

Once the raw images are cropped to a circle shape, the rotation of each image is smooth and easy to implement. Here, we provide a group of raw images alongside a group of images after feature alignment in Figure 5. It is clear that after the 'crop and rotate' procedure, the feature of the drone-view image is aligned. It is worth mentioning that, in real- world test stage, although there is no index to show which angle we should take to rotate the captured image, the orientation can be easily obtained without any GPS information. Instead, a compass is enough.  

Once the feature alignment is complete, the aligned drone-view images boost the feature extraction performance of the CNN. As we can see in the experiment section, even with the vanilla model, the performance is significantly improved with no modifications. 

There is another pipeline of aligned feature extraction based on aligned partial features called part-based feature CNN \cite{sun2018beyond}. This method usually conducts uniform partitions on the conv-layer for learning partial features. However, it does not explicitly partition the images; instead, it takes a whole image as the input and outputs a convolutional feature. Thus, the architecture of partial feature extraction network is concise, with slight modifications on the backbone network. 

In line with the above, we introduce an additional branch of the vanilla approach. In departure from \cite{sun2018beyond}, we do not abandon the global feature branch as although there is no feature alignment procedure that can be done for the street-view images or additional Google pictures, the additional information provided by these image sources is still useful for providing guidance for common feature space generation.

We mainly adopt two partition methods to get the partial feature: the regular partition and dense partition. Considering the equal importance of the target surroundings, we treat each partition equally and apply no coefficient. In section 4.3.2, we will demonstrate the performance of global feature under the guidance of partial features is greatly improved when compared to the vanilla method. We will also discuss different partition strategies to analyze which one is better for extracting the partial features.

\section{Experiment}

\subsection{Dataset}
University-1652 is a recently released dataset for multi-view multi-source geo-localization tasks. This dataset cover 1652 architectures of 72 universities around the world as target locations. Thus, there is 1,652 classes of different source images, including drone-view, satellite, ground-view  and common-view images. In total, there are 5,580 street-view images and 21,099 common-view images from Google Map and Google Images, respectively. The images collected from Google Images only serve as an extra training set, not a test set. Every building has one satellite-view image on average. The images were cropped from the drone-video every 15 frames, which is around 20 degrees, resulting in 54 drone-view images. Overall, every building has a total of 58.38 reference images. If using extra Google-retrieved data, there will be 16,644 ground view images per building for training. There are 701 classes with 50,218 images in the training set and 951 classes with 51,355 drone-view images and 793 classes with 2,921 ground-view images in the gallery set, including 701 classes of drone-view images in the query set. There is no overlap between the 33 universities in the training set and 39 universities in  test sets.

\subsection{Implementation Details}

We implement the model on the ResNet50\cite{he2016deep} backbone with several optimizations relative to the original one in University-1652. We add an additional branch before the global average pooling (GAP) layer. As illustrated in Figure 4, we maintain the vanilla branch extracting global feature with no modifications. Moreover, we have one or more copies of 3D tensor which represents features after layer 4 of the ResNet50. For example, for the '3+3' partition in Table3, we partition the 3D tensor on both vertical and horizontal dimensions to get six equal parts along both the height and width of the 3D tensor. After dividing the 3D tensor, we use average pooling to average all the column vectors in the same stripe into a single part-level vector.  We then employs a convolutional layer to reduce the vectors to a 512-dim vector; finally, we use a classifier implemented with a fully connected (FC) layer and follow the softmax function to predict the identity of the input. During testing, we employ the 2048-dim feature rather than the 512-dim feature to compute similarity, as the experiment shows that 512-dim features are lower than 2048-dim features on accuracy metrics.

\begin{figure}[!b]
\begin{center}
   \includegraphics[width=0.9\linewidth]{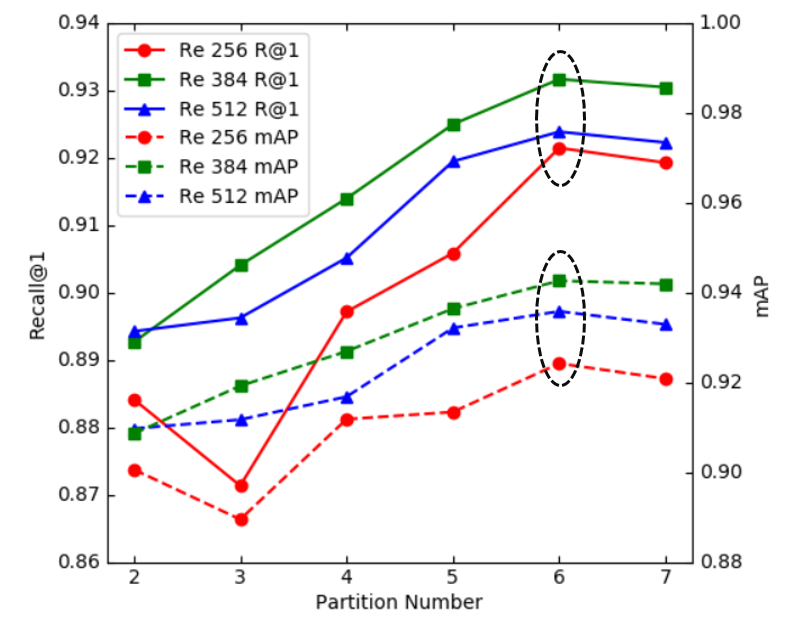}
\end{center}
   \caption{Recall@1 and mAP with different input sizes and partition numbers using regular partitioning. Here, we only show the drone-satellite branch. The black dashed circle indicates the highest accuracy.   }
\label{fig:long}
\label{fig:onecol}
\end{figure}

\subsection{Ablation Study}

\begin{figure}[!b]
\begin{center}
   \includegraphics[width=1.0\linewidth]{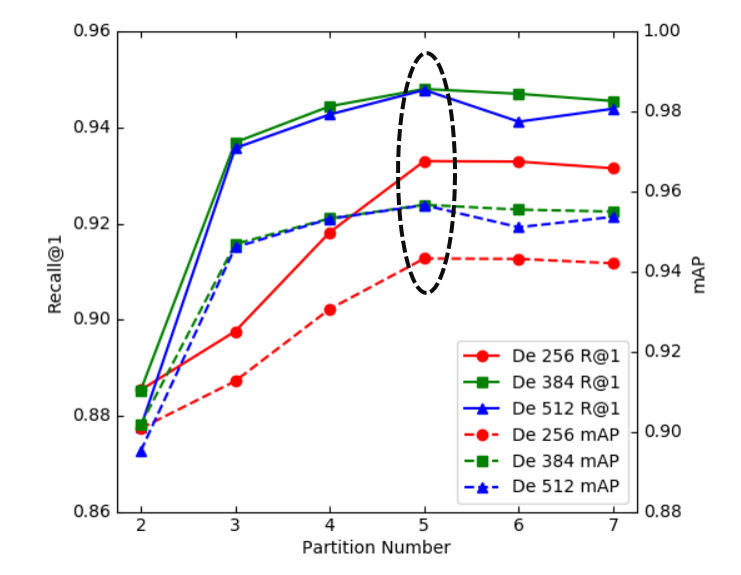}
\end{center}
   \caption{Recall@1 and mAP with different input sizes and partition numbers using dense partitioning. Here, we only show the drone-satellite branch. The black dashed circle indicates the highest accuracy. }
\label{fig:long}
\label{fig:onecol}
\end{figure}

\subsubsection{Crop, Rotate and Style Alignment}

\begin{table}[]
    \centering
\begin{tabular}{c|ccccccc}
\hline  %添加表格头部粗线
 Method & R@1 & R@5 & R@10 & mAP \\
\hline  %添加表格中横线

vanilla & 58.49 & 78.67 & 85.23 & 63.13  \\
vanilla+C & 60.59 & 82.73 & 87.51 & 66.13   \\
vanilla+C+R & 70.03 & 86.94 & 91.18 & 73.86  \\
vanilla+C+R+A & \textbf{71.70} & \textbf{88.79} & \textbf{93.23} & \textbf{76.12} \\

\hline %添加表格底部粗线
\end{tabular}
    \caption{We evaluate the different data preprocessing methods. The basic method is from University-1652 using instance loss. C represents circle crop, R represents rotation and A represents style alignment. Note that we here use only the global feature after the bottleneck of linear block to test the model performance and we only show the drone-satellite branch.}
    \label{tab:my_label}
\end{table}

Table 1 shows the ablation study results of our data preprocessing method. We add 'crop', 'rotate' and 'style alignment' consequently on the raw data. Note that we here adopt 512-dim feature to compare with the model performance on raw data. We make no modifications to the vanilla model using only instance loss. We still take Google images and street-view images as an extra source to guide model using the same method as before. The experimental results show that all three data preprocessing methods gain significant performance improvement. The most effective method is 'rotate', which gain about 10 percent improvement on Recall@1 and 7 percent improvement on mAP.

\subsubsection{Partial Feature Guidance}

\begin{table}[!b]
    \centering
\begin{tabular}{cc|cccccc}
\hline  %添加表格头部粗线
  Method & size & R@1 & R@5 & R@10 & mAP \\
\hline  %添加表格中横线

basic & 256 & 70.03 & 86.94 & 91.18 & 73.86  \\
basic & 384& 70.04 & 86.70 & 90.71 & 73.82   \\
basic & 512 & 66.62 & 86.39 & 90.59 & 71.06  \\
basic+P & 256 & 87.14 & 95.08 & 96.61 & 88.96  \\
basic+P & 384 & \textbf{90.41} & \textbf{97.13} & \textbf{98.15} & \textbf{91.93}   \\
basic+P & 512 & 89.63 & 96.33 & 97.62 & 91.18  \\

\hline %添加表格底部粗线
\end{tabular}
    \caption{We evaluate the performance improvement resulting from feature extraction on different input sizes. The basic method contains our data preprocessing method including 'crop' and 'rotate' without 'style alignment' as we treat the style alignment method as an independent process in the testing stage. 'P' stands for partial feature, which is the only difference from the basic method. Here, we choose 256, 384 and 512 as the input side length of images considering the raw image size (512x512). The partition strategy is fixed to be '3+3' regular partition which divides the 3D tensor equally into 6 parts both vertically and horizontally. Note that we  use only the feature vector with 2048-dim in the basic+P group to test the model performance, and we only show the drone-satellite performance.}
    \label{tab:my_label}
\end{table}

We evaluate the performance improvement with partial feature extraction on different input sizes. As we can see from table 2, employing partial feature in the basic global feature model significantly improves the model performance on accuracy. All three input sizes achieved significant improvement. Input size 384 shows the highest accuracy on R@1 R@5 and R@10 under '3+3' partition strategy. Input size 512 shows the highest accuracy on mAP under this strategy. The result indicates that input size is an important factor on model performance under same partition strategy. We will discuss this in next section.

\begin{figure*}[!h]
\begin{center}
   \includegraphics[width=0.9\linewidth]{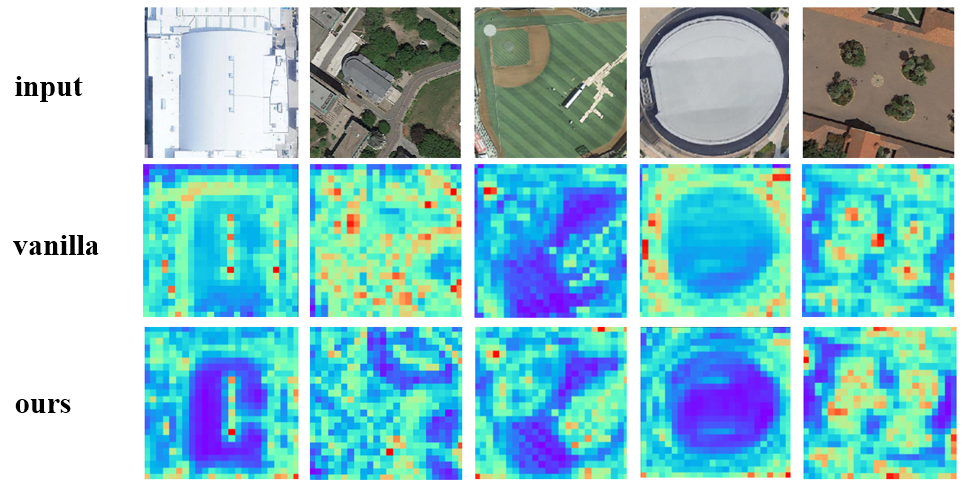}
\end{center}
   \caption{Embedding visualizations. Here we adopt \cite{zheng2017discriminatively} to visualize the CNN embeddings. The heatmap of our method is clearer than the vanilla method. Moreover, the key points of the target building and its surroundings are more specific and accurate, and we can easily find the segment border.  }
\label{fig:long}
\label{fig:onecol}
\end{figure*}

\subsection{Partition Strategy and More}

\begin{table}[!b]
    \centering
\begin{tabular}{c|cc|cc}
\hline  %添加表格头部粗线
{\multirow{2}*{strategy}} &  \multicolumn{2}{c|}{drone-satellite} & \multicolumn{2}{c}{satellite-drone}\\
~&  R@1 & mAP & R@1 & mAP\\
\hline  %添加表格头部粗线

V+CL &52.39& 57.44 & 63.91 & 52.24 \\
V+TL & 55.18& 59.97 & 63.62 & 53.85   \\
V+IL & 58.23 & 62.91 & 74.47 & 59.45 \\
\hline  %添加表格头部粗线
Ours+Re &93.17 & 94.27 & 96.86 & 90.96   \\
Ours+Re+A &93.21 & 94.38 & 97.05 & 91.23   \\
Ours+De & 94.78 & 95.67 & 98.15 & 93.74   \\
Ours+De+A & \textbf{94.84} & \textbf{95.80} & \textbf{98.35} & \textbf{94.02}   \\

\hline %添加表格底部粗线
\end{tabular}
    \caption{Comparison to the existing methods on University-1652. 'V'  represents the vanilla method, while 'CL','TL' and 'IL' represent contrastive loss, triplet loss and instance loss respectively. Our full model with style alignment strategy 'A' can bring the drone-view-based geo-localization task into practical use.}
    \label{tab:my_label}
\end{table}

\begin{table*}[]
    \centering
\begin{tabular}{c|c|c|cccc|cccc}
\hline  %添加表格头部粗线
{\multirow{2}{*}{Group}} & {\multirow{2}*{input size}} & {\multirow{2}*{strategy}} &  \multicolumn{4}{c|}{drone-satellite} & \multicolumn{4}{c}{satellite-drone}\\
~& ~& ~& R@1 & R@5 & R@10 & mAP & R@1 & R@5 & R@10 & mAP\\
\hline  %添加表格中横线

{\multirow{18}{*}{Re-Partition}}& {\multirow{6}*{256}} &2+2 & 88.42 & 95.65 & 97.04 & 90.07 & 94.29 & 96.01 & 96.86 & 86.52\\
~& ~ & 3+3 & 87.14 & 95.08 & 96.61 & 88.96 & 94.15 & 96.43 & 97.00 & 85.53 \\
~& ~ & 4+4 & 89.72 & 96.27 & 97.54 & 91.19 & 94.29 & 96.29 & 97.43 & 86.86 \\
~& ~ & 5+5 & 90.59 & 96.57 & 97.64 & 91.35 & 93.58 & 96.58 & 97.29 & 87.65 \\
~& ~ & 6+6 & \textbf{92.15} & \textbf{96.73} & 97.77 & 92.43 & \textbf{95.15} & \textbf{97.43} & \textbf{98.00} & \textbf{89.04} \\
~& ~ & 7+7 & 91.93 & 96.68 & \textbf{97.84} & \textbf{93.09} & 94.29 & 97.29 & 97.72 & 88.59 \\
\cline{2-11}
~& {\multirow{6}*{384}} & 2+2 & 89.27 & 96.34 & 97.50 & 90.87 & 95.44 & 97.43 & 97.72 & 88.41\\
~& ~ & 3+3 & 90.41 & 97.13 & 98.15 & 91.93 & 96.01 & 97.29 & 98.15 & 88.13 \\
~& ~ & 4+4 & 91.40 & 97.05 & 98.09 & 92.70 & 95.58 & 97.57 & 98.43 & 88.22 \\
~& ~ & 5+5 & 92.50 & 97.51 & 98.34 & 93.65 & 95.86 & 97.72 & 98.00 & 90.42 \\
~& ~ & 6+6 & \textbf{93.17} & 97.97 & 98.69 & \textbf{94.27} & 96.58 & 98.29 & 98.57 & \textbf{90.96} \\
~& ~ & 7+7 & 93.05 & \textbf{98.06} & \textbf{98.79} & 94.20 & \textbf{96.72} & \textbf{98.86} & \textbf{99.14} & 90.32 \\
\cline{2-11}
~& {\multirow{6}*{512}} &2+2 & 89.43 & 96.33 & 97.54 & 90.98 & 94.86 & 97.00 & 97.43 & 87.62\\
~& ~ & 3+3 & 89.63 & 96.33 & 97.62 & 91.18 & 94.01 & 96.72 & 97.57 & 87.29 \\
~& ~ & 4+4 & 90.52 & 96.86 & 97.88 & 91.69 & 95.44 & 97.29 & 98.15 & 87.94\\

~& ~ & 5+5 & 91.95 & 97.51 & 98.36 & 93.22 & 96.72 & 98.15 & 98.29 & 89.20\\
~& ~ & 6+6 & \textbf{92.39} & 97.50 & 98.55 & \textbf{93.59} &95.86 & 97.72 & 98.15 & 89.05\\
~& ~ & 7+7 & 92.23 & \textbf{97.73} & \textbf{98.69} & 93.30 & \textbf{96.86} & \textbf{98.43} & \textbf{98.86} & \textbf{90.75}\\

\hline  %添加表格中横线

{\multirow{15}*{De-Partition}}& {\multirow{5}*{256}} &2x2 & 88.54 & 95.21 & 96.63 & 90.09 & 95.72 & 97.15 & 98.29 & 87.61\\
~& ~ & 3x3 & 89.76 & 96.21 & 97.32 & 91.27 & 94.29 & 96.43 & 97.00 & 88.10 \\
~& ~ & 4x4 & 91.82 & 97.32 & 98.30 & 93.07 & 96.01 & 97.72 & 98.29 & 90.81 \\
~& ~ & 5x5 & \textbf{93.30} & \textbf{97.82} & \textbf{98.56} & \textbf{94.33} & 96.01 & 97.72 & 98.43 & 91.41 \\
~& ~ & 6x6 & 93.29 & 97.72 & 98.54 & 94.32 & \textbf{96.15} & \textbf{97.86} & \textbf{98.57} & \textbf{91.53} \\
\cline{2-11}
~& {\multirow{5}*{384}} &2x2 & 88.53 & 95.76 & 97.09 & 90.18 & 94.86 & 98.29 & 98.57 & 86.92\\
~& ~ & 3x3 & 93.69 & 98.07 & 98.64 & 94.69 & 97.29 & 98.57 & 99.14 & 92.53\\
~& ~ & 4x4 & 94.44 & 98.24 & 98.79 & 95.33 & 97.15 & 98.57 & 98.86 & 92.77\\
~& ~ & 5x5 & \textbf{94.80} & \textbf{98.61} & \textbf{99.25} & \textbf{95.67} & \textbf{97.57} & 99.14 & \textbf{99.57} & 93.50 \\
~& ~ & 6x6 & 94.70 & 98.41 & 99.17 & 95.55 &\textbf{97.57} & \textbf{99.29} & \textbf{99.57} & \textbf{93.74} \\
\cline{2-11}
~& {\multirow{5}*{512}} &2x2 & 87.76 & 95.41 & 96.85 & 89.52 & 93.72 & 97.15 & 98.15 & 85.93\\
~& ~ & 3x3 & 93.57 & 98.11 & 98.76 & 94.61 & 97.15 & 98.72 & 99.43 & 91.48\\
~& ~ & 4x4 & 94.27 & 98.34 & 98.89 & 95.21 & 96.72 & 98.29 & 98.86 & 92.47\\
~& ~ & 5x5 & \textbf{94.78} & \textbf{98.63} & \textbf{99.12} & \textbf{95.66} & 97.00 & 98.72 & 99.14 & 92.93 \\
~& ~ & 6x6 & 94.12 & 98.45 & 99.11 & 95.11 & \textbf{98.15} & \textbf{99.29} & \textbf{99.57} & \textbf{93.24} \\

\hline %添加表格底部粗线
\end{tabular}
    \caption{Automatic Evaluation Score. `Re' represents `regular' and `De' represents `dense'.}
    \label{tab:my_label}
\end{table*}

The use of different partition strategies results in different model performance. Considering the equal importance of all surroundings, the partitioning should be symmetrical. We adopt two groups  partition strategies with different numbers of parts to evaluate their relative effectiveness. The first group is the regular partition group. In this group, we equally divide the 3D tensor into n equal parts both vertically and horizontally to get ${2\times n}$ parts of vectors. The second group is the dense partition group; here we divide the 3D tensor into ${n}^2$ parts, as illustrated in Figure 6. Note that the dense partition group only needs one copy of the 3D tensor from global feature extraction branch while regular partition group needs two copies. In table 4, we show the different partition strategies for these two groups with different input sizes and partition numbers.

The results for the different partition strategies in table 3 indicates two main factors to model performance, which are partition number and input size. Here we have three input size including 256, 384, and 512; these equate to 1/2, 3/4 and equal to the raw image side length respectively. In each group, we apply several different partition numbers. We first test our model performance using regular partitioning. The model achieves the best accuracy when using a 6+6 partitioning strategy with 384 or 512 input size (figure 7). We then apply dense partitioning to train the model. In this partition case, the model achieves the best accuracy when using a 5x5 partition strategy (figure 8). Meanwhile, it is interesting to see that the dense partition model gains significant improvement when the partition number goes up. By contrast, the improvement under regular partitioning increases relatively slowly. We can guess that the reason lies in how 'independent' the part feature is. Considering the overlap between different feature vectors of the regular partition, the ability to cover more features of the surroundings will be weakened, leading to lower performance when representing partial features in the testing stage compared to the dense partition strategy.

\subsection{Performance Evaluation}

Here, we present the visualization heatmap in Figure 9, while the best performance between the vanilla method, the basic method with all data pre-processing methods applied and the full model using different partition strategies are represented in Table 3. Compared to the vanilla method with instance loss, our approach is far beyond the baseline, with about 40 percent improvement on both Recall@1 and mAP. Figure 9 shows the heatmap for the vanilla method and our full approach, which reveals that our approach is more robust and covers more surroundings of the target building to assist feature representation.

\section{Conclusion}

%In this paper, we have proposed three strategies, namely color uniforming, feature alignment and partial feature extraction, for application to the drone-based multi-view geo-localization task. Experiments have demonstrated that these strategies significantly improve the geo-localization accuracy. The proposed method will be able to accelerate the practical application of drone navigation and localization.

In this paper, we have proposed style and spatial alignment approaches for multi-view drone-based geo-localization. Specifically, we have proposed an elegant orientation-based method to align the patterns and introduced a new branch to extract aligned partial feature. In addition, we have provided a style alignment strategy to reduce the variance in image style and enhance the feature unification. We have verified the effectiveness of the proposed approach on the large-scale benchmark dataset. Besides, we have conducted ablation studies to confirm the influence of each component. From the experimental results, we observe that all the components contribute significantly to the overall improvement.

\bibliographystyle{ACM-Reference-Format}
\bibliography{sample-base}

% \appendix









\end{document}